%% file: main.tex
\documentclass{article}
\usepackage{spconf,amsmath,epsfig}

\usepackage{times}  
\usepackage{helvet}  
\usepackage{courier}  
\usepackage[hyphens]{url}  
\usepackage{graphicx} 
\usepackage{caption} 

\usepackage{algorithm}
\usepackage{algorithmic}

\usepackage[utf8]{inputenc} 
\usepackage{booktabs}       
\usepackage{amsfonts}       
\usepackage{nicefrac}       
\usepackage{microtype}      
\usepackage{xcolor}         
\usepackage{comment}
\usepackage{array}
\usepackage{amsmath} 
\usepackage{multirow}
\usepackage{enumitem}
\usepackage{makecell}
\usepackage{xfrac}
\usepackage{soul}
\usepackage{subfigure}

\let\OLDthebibliography\thebibliography
\renewcommand\thebibliography[1]{
  \OLDthebibliography{#1}
  \setlength{\parskip}{0pt}
  \setlength{\itemsep}{0pt plus 0.3ex}
}

\begin{document}\sloppy
\def\x{{\mathbf x}}
\def\L{{\cal L}}

\title{Compact Real-time Radiance Fields with Neural Codebook}
%
\name{Lingzhi Li, Zhongshu Wang, Zhen Shen, Li Shen, Ping Tan}
\address{Alibaba Group, Beijing, China}
\address{\tt\small \{llz273714, zhongshu.wzs, zackary.sz\}@alibaba-inc.com, lshen.lsh@gmail.com,  pingtan@ust.hk}

\maketitle

\begin{abstract}
Reconstructing neural radiance fields with explicit volumetric representations, demonstrated by Plenoxels, has shown remarkable advantages on training and rendering efficiency, while grid-based representations typically induce considerable overhead for storage and transmission. In this work, we present a simple and effective framework for pursuing compact radiance fields from the perspective of compression methodology. By exploiting intrinsic properties exhibiting in grid models, a non-uniform compression stem is developed to significantly reduce model complexity and a novel parameterized module, named Neural Codebook, is introduced for better encoding high-frequency details specific to per-scene models via a fast optimization. Our approach can achieve over 40 $\times$ reduction on grid model storage with competitive rendering quality. In addition, the method can achieve real-time rendering speed with 180 fps, realizing significant advantage on storage cost compared to real-time rendering methods. 
\end{abstract}

\input{intro}
\input{related_work}

\input{alg}

\input{exp}

\section{Conclusion}

We propose a compression framework to significantly reduce the storage cost of voxel grid models by the use of non-uniform compression. The introduction of Neural Codebooks
enables effective encoding high-frequency information to better preserve scene-specific details. Experiments on two widely-used datasets demonstrate remarkable advantage of our method on storage cost compared to existing real-time rendering methods.


{
\bibliographystyle{IEEEbib}
\bibliography{egbib}
}

\clearpage
\input{supp}

\end{document}

%% file: intro.tex
\section{Introduction}
Neural representations have recently emerged as a promising direction for reconstructing shapes \cite{MateuszMichalkiewicz2019ImplicitSR}, lighting \cite{PratulPSrinivasan2020NeRVNR} and radiance fields \cite{NeRF,NeRF++} of scenes with intricate geometry and appearance. 
Neural radiance field methods \cite{NeRF, NeRF++} have shown compelling quality for realizing free-view photo-realist rendering. By means of volumetric rendering techniques, a continuous mapping function realized with multi-layer perceptron (MLP) is learned to map 5D coordinates (3D position with a viewing direction) to view-dependent colors and volume densities, so pixel colors are rendered by accumulating sampling points along casting rays. However, these methods suffer from costly training and rendering issue due to the need of tremendous sampling, each of which passes through large networks. 

\begin{figure}[t]
\centering
\includegraphics[width=1.0\linewidth]{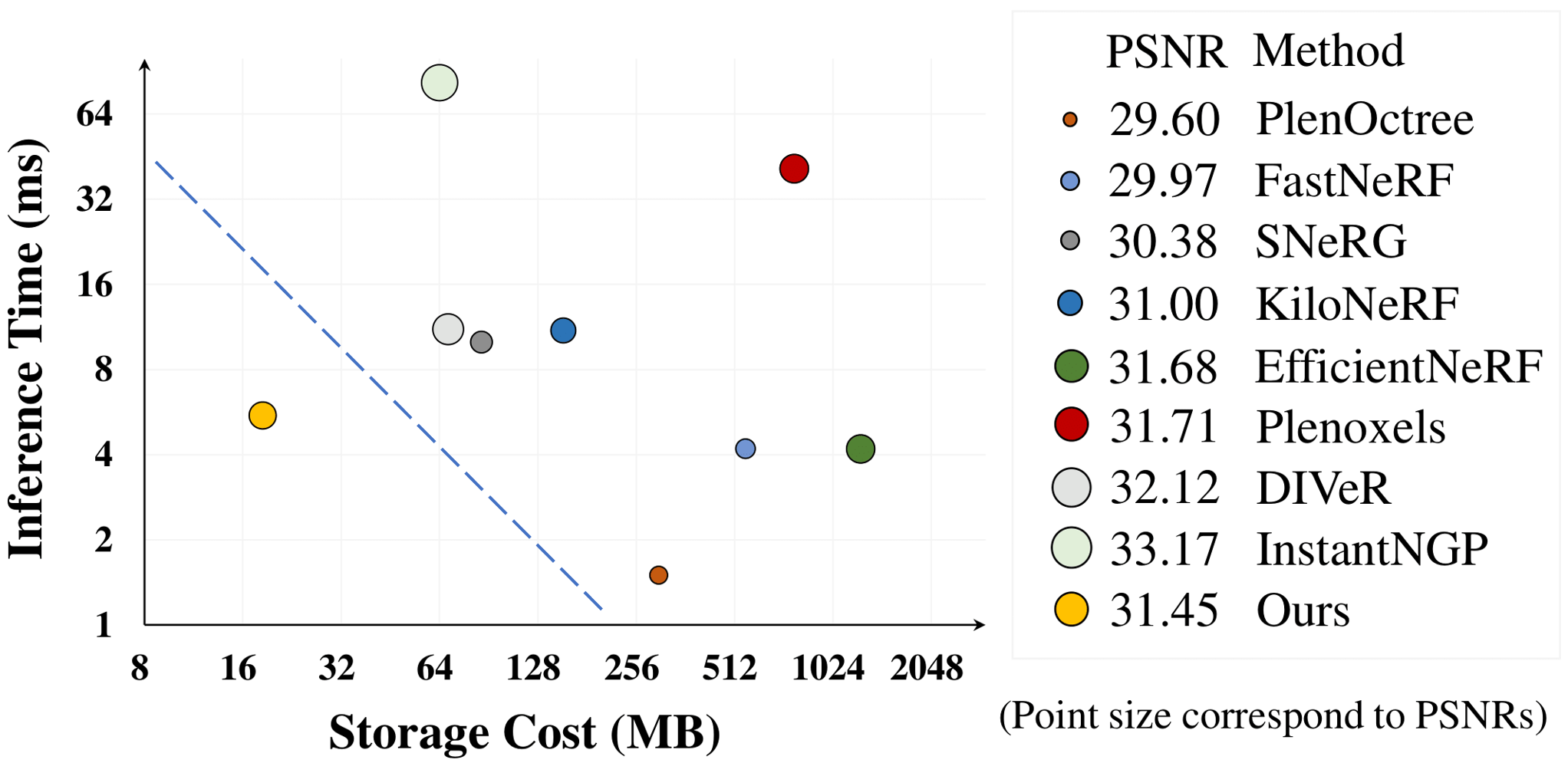}
\caption{Quantitative Results on Synthetic-NeRF for comparing real-time rendering methods in terms of storage cost (MB), inference time (ms) and rendering quality (PSNR).}
\label{fig:compare}
\vspace{-1em}
\end{figure}

Recent advances introduce discretized grid structures into radiance field reconstruction \cite{yu2021plenoxels, PlenOctrees, sun2021direct_dvgo, sparsevoxel} and have shown that representing a scene with volume grid has significant advantages on inference efficiency, i.e., reaching interactive speed or even real-time rendering speed. Unlike relying on network evaluation in implicit methods, color features and volume densities are explicitly stored in the voxels of a grid model.   
However efficiency gains are at the price of storage allocation, typically costing hundreds of megabytes even more for representing a single scene. As shown in Figure~\ref{fig:compare}, most of real-time inference methods cost large storage, which is prohibitive for storage and transmission in real-world applications.   

In this paper we aim to address the storage issue induced by using voxel grids, aiming to realize compact real-time radiance fields with competitive rendering quality. We consider it from a perspective of compression methodology, i.e., reducing the storage cost through an efficient compression phase and recovering the grid model through {\it one-time} deployment (decompression). We conduct empirical analysis and propose a simple yet efficient compression framework based on several practical guidance, for better leveraging the intrinsic properties existing in volume grid models. The framework is illustrated in Fig.~\ref{fig:NCB framework}. Specially, 
we introduce a non-uniform compression strategy to compress a huge full-scale grid model to reach significantly lower storage overhead, and exploit a position embedding modulated component, i.e., Neural Codebook, to effectively encode high 
frequency information specific to each grid model in order to better preserve the scene-specific details. Our method can realize about $40\times$ storage reduction compared to original grid models \cite{yu2021plenoxels} with competitive quality (less than 0.3dB drop) on benchmark datasets.  
We further exploit several strategies to improve inference efficiency without the need of octree-based structures \cite{PlenOctrees}. We are able to realize real-time rendering with 180 fps, achieving 
smaller storage cost than modern real-time rendering methods (in Fig.~\ref{fig:compare}) and showing the best tradeoff between storage cost and rendering speed.

%% file: related_work.tex
\section{Related Work} 

A series of strategies are employed to improve the training and inference efficiency of Neural Radiance Fields (NeRF) \cite{NeRF}. 
FastNeRF \cite{fastnerf} accelerates rendering speed by efficiently caching and querying radiance maps. KiloNeRF \cite{kilonerf} reduces processing time by representing scenes with a lot of small MLPs.
\cite{MartinPiala2021TermiNeRFRT,ThomasNeff2021DONeRFTR} 
introduce auxiliary networks to predict sampling points locations  likely placed around surface. \cite{RuizhiShao2021DoubleFieldBT} 
jointly reconstructs surfaces/meshes with radiance fields . 

A family of methods achieve fast or even real-time rendering by storing features in a voxel grid structure. The features are either converted from a pretrained NeRFs for inference only \cite{PlenOctrees, SNeRG} or learned in a hybrid manner with small MLPs \cite{sun2021direct_dvgo, sparsevoxel, InstantNGP}. Plenoxels leverages spherical harmonics to realize view-dependent appearance without neural networks. However, these methods typically take huge storage cost, even two orders of magnitude larger than implicit methods. TensorRF \cite{chen2022tensorf} 
factorizes voxel grids to achieve compact model sizes 
however obviously slowing inference speed (requiring 200ms for rendering 800 × 800). 
Our method aims to address the storage issue of explicit grid and  achieve real-time inference. 

Our work is also related to model compression, which aims to reduce model size of large networks while remaining accuracy, typically through the techniques of parameter pruning \cite{han2015deep},  network quantization 
\cite{jacob2018quantization}, low-rank approximation \cite{jaderberg2014speeding} and knowledge distillation \cite{hinton2015distilling}. 
The sparsification operation in \cite{yu2021plenoxels, PlenOctrees} can be treated as a kind of parameter pruning. \cite{takikawa2022variable, lingzhi_vqrf} introduce codebooks
for quantizing feature vectors and mapping into feature index.
TensorRF \cite{chen2022tensorf} is built upon low-rank approximation. Our method can be seen as a combination of volume data compression \cite{nguyen2001rapid} and knowledge distillation with domain priors.

%% file: alg.tex

\begin{figure*}[t]  
    \centering
    \includegraphics[width=0.8\linewidth]{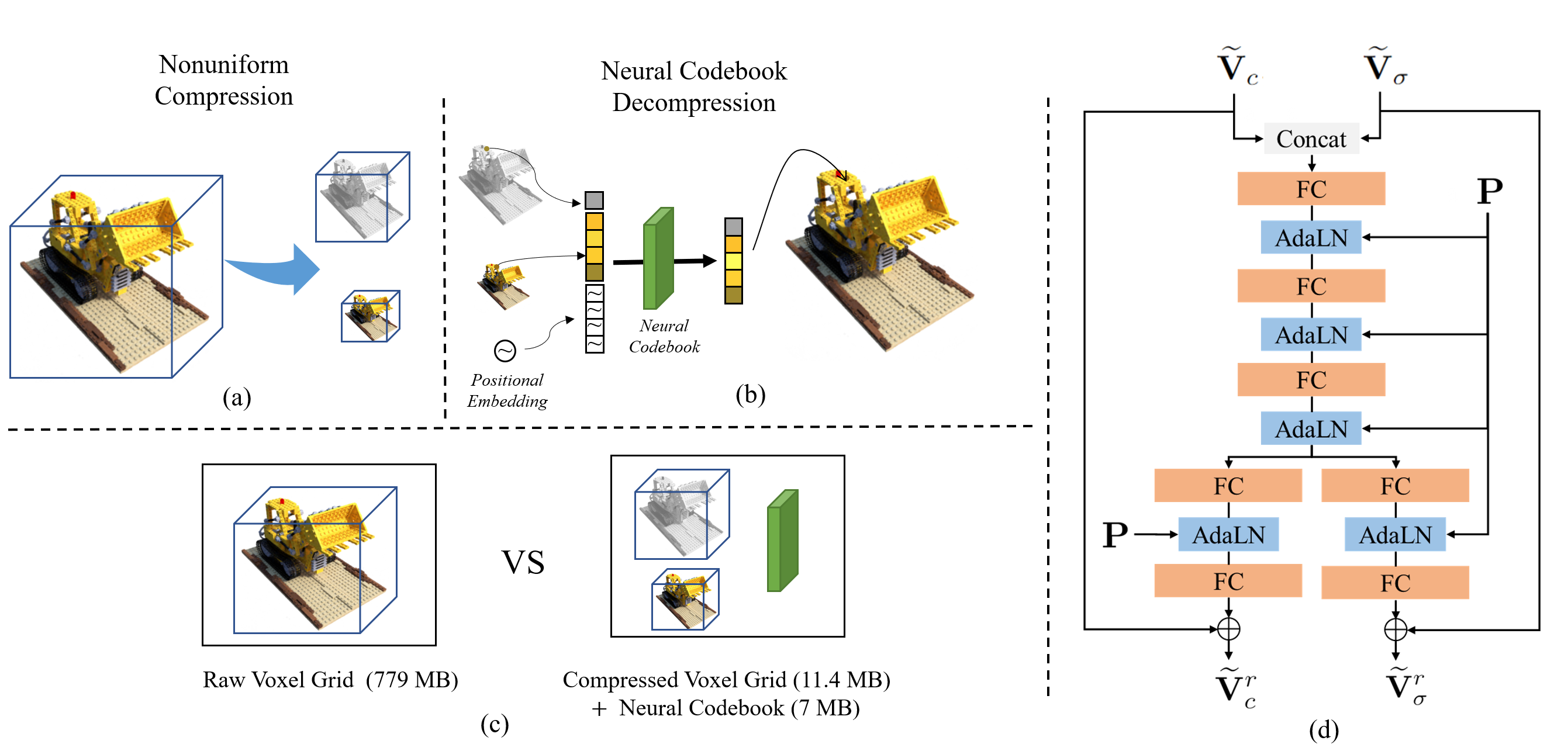}
    \caption{The illustration of the framework. (a-c) Given an explicit grid model, we effectively compress it via nonuniform compression and a learnable Neural Codebook (NCB) for preserving details. The grid model is restored by performing one-time decompression without loss on rendering efficiency given novel views.  
    (d) Illustration of NCB architecture.} 
    \label{fig:NCB framework}  
\end{figure*}

\section{Method}

\subsection{Overview}

We propose a compression framework to address the storage issue of voxel grid models, including compressing grid models to a much smaller size through {\it Nonuniform Compression} and learning a parameterized module {\it Neural Codebook} for details recovery, by employing intrinsic properties in grid models.

Formally, a grid model \cite{yu2021plenoxels} is comprised of volume density and the features inducing view-variant colors, denoted as $\mathbf{V}_{\sigma} \in \mathbb{R}^{H\times W\times K}$ and $\mathbf{V}_{c} \in \mathbb{R}^{H\times W\times K\times C}$. Here $H$, $W$ and $K$ denote spatial resolutions. Color features are formed by the coefficients of a set of spherical harmonics (SH) and the dimension is $C$. The mode is optimized via volume rendering technique, i.e., casting a batch of rays $\mathbf{r}$ from camera center, and sampling $N$ points along each ray to estimate the corresponding colors $\hat{C}(\mathbf{r})$ according to 
\begin{equation}\label{eq:render} 
    \widehat{C}(\mathbf{r}) = \sum_{i=1}^N T_i\cdot \alpha_i \cdot \mathbf{c}_i,
\end{equation}
\begin{equation}\label{eq:alpha_t}
    \alpha_i = 1 - \exp(-\sigma_i \delta_i),\quad T_i = \prod_{j=1}^{i-1} (1 - \alpha_j),
\end{equation}
where $\sigma_i$ and $\mathbf{c}_i$ denote the density and color at point $i$, calculated via tri-linear interpolation  
with neighboring voxels. 
The model size of grids grows cubically with respect to spatial dimension. When initializing a grid with $512^3$, the memory cost is about $13$ GB. Although the overall model size can decrease to $~820$ MB via the traditional compression operations of sparsification and conversion to float16 type as in \cite{PlenOctrees, yu2021plenoxels}, such a size is still less satisfied for storage and transmission in real-world applications.

We address the storage issue from the perspective of compression methodology, i.e., reducing the resources required to store and transmit grid models via a compression phase, and restoring them from the encoded models \textit{at once} at the decompression phase. 
The insights behind the method are derived from the following  practical guidance (which will be elaborated in the following sections):
\vspace{-1mm}
\begin{itemize}
    \item Compared to higher-dimensional color features, density information influences more for preserving model ability. Compression color features with a higher rate is more effective.
    \vspace{-1mm}
    \item  The voxels with large importance scores only occupy a fraction, but they contribute much to rendering quality.
    \vspace{-1mm}
    \item Better employing the correlation between color features and density is helpful for fine details recovery.   
\end{itemize}
\vspace{-1mm}
A full-scale model can be gotten by one-time deployment(decompression) with comparable deployment time to the original model and no loss for rendering speed.

As an orthogonal improvement, we introduce engineering strategies to accelerate rendering without the need of converting to an octree structure which may sacrifice rendering quality. Specially, we 1) combine all cuda threads, 2) double sampling steps in ray casting, 3) terminate rays when the accumulated density reaches 99\%, 4) prefecth all data-pointer before trilinear interpolation. Detailed descriptions can refer to supplemental material. We can finally achieve rendering with 180 fps, reducing the cost from $41$ ms/frame to $5.5$ ms/frame.

\begin{table}[h]
    \caption{
    The effect of downsampling with different scales on color features and volume density. 
    }
    \vspace{-0.5mm}
    \label{tab:Nonuniform downsample}
    \centering
    \small
    \begin{tabular}{l@{\hskip 8pt}@{\hskip 5.5pt}|c@{\hskip 5.5pt}c@{\hskip 5.5pt}c@{\hskip 5.5pt}|@{\hskip 5.5pt}c@{\hskip 5.5pt}c@{\hskip 5.5pt}c@{\hskip 5.5pt}}
    \hline
    \toprule
    & PSNR$\uparrow$ & SSIM$\uparrow$ & LPIPS$\downarrow$  & Size$\downarrow$  \\
    \midrule
    $\downarrow^{\sfrac{1}{2}} (\mathbf{V}_c)$ + $\downarrow^{\sfrac{1}{2}} (\mathbf{V}_{\sigma})$             & 26.55 & 0.920 & 0.095 &  27.9MB         \\
    \hspace{0.4cm} + codebook & 31.52 & 0.950 & 0.058 & 34.4MB \\
    \midrule
    $\downarrow^{\sfrac{1}{2}} (\mathbf{V}_c)$ + $\downarrow^{\sfrac{1}{4}} (\mathbf{V}_{\sigma})$            & 23.11 & 0.884 & 0.144  &   26.3MB       \\
    \hspace{0.4cm} + codebook & 31.52 & 0.956 & 0.062 &    33.8MB \\
    \midrule
    $\downarrow^{\sfrac{1}{4}} (\mathbf{V}_c)$ + $\downarrow^{\sfrac{1}{2}} (\mathbf{V}_{\sigma})$            & 25.20 & 0.897 & 0.118 &  6.2MB    \\
    \hspace{0.4cm} + codebook & 31.45 & 0.955 & 0.058 & 13.7MB \\
    \midrule
    $\downarrow^{\sfrac{1}{4}} (\mathbf{V}_c)$ + $\downarrow^{\sfrac{1}{4}} (\mathbf{V}_{\sigma})$            & 22.67 & 0.872  & 0.1569 &   4.6MB   \\
    \hspace{0.4cm} + codebook & 31.43 & 0.954  &  0.070 &   12.1MB \\
    \bottomrule
    \end{tabular}
    \vspace{-1em}
\end{table}
\subsection{Nonuniform Compression on Grid Models}

Given a well-trained grid model $\mathcal{V}=\{\mathbf{V}_{\sigma}, \mathbf{V}_c\}$,
 a much smaller model can be intuitively obtained by employing a down-sampling operation on the grid along with resolution. As the color features $\mathbf{V}_c$ is typically tens of times larger than the volume density $\mathbf{V}_{\sigma}$, we adopt different down-sampling scales on $\mathbf{V}_c$ and $\mathbf{V}_{\sigma}$,
\begin{equation}\label{eq:downsample}
\widehat{\mathbf{V}}_c = \downarrow^{S_c} (\mathbf{V}_c),\quad \widehat{\mathbf{V}}_{\sigma} = \downarrow^{S_\sigma} (\mathbf{V}_\sigma),
\end{equation}
where $S_c$ and $S_\sigma$ denote  down-sampling ratios for color features and density.
Intuitively, the compressed model $\{\widehat{\mathbf{V}}_{\sigma}, \widehat{\mathbf{V}}_c\}$ can be directly up-sampled to the size of the original grid during decompressing through tri-linear interpolation, a parameter-free manner, to obtain the recovered grid model, $\{\widetilde{\mathbf{V}}_{\sigma}, \widetilde{\mathbf{V}}_c\}$.

We evaluate these models to assess the effect for rendering quality (shown in Table \ref{tab:Nonuniform downsample}). We can observe that density information contributes more than color features for preserving model rendering quality. Using a smaller down-sampling rate on volume density would obviously reduce model capacity while contributing less for storage saving. In contrast, using a smaller down-sampling rate on color features can obtain a model with much lower storage, while with moderate degeneration on rendering quality. It is more efficient to use a larger compression rate on color features.

We further define an importance score for each voxel as follows. When sampling a point $\mathbf{x}_i$, the density $\sigma_i$ is calculated by linearly interpolating the $8$ adjacent voxels $\mathbf{v}_l \in \mathcal{N}_i$ according to the distance between them, where the linear weight of $\mathbf{v}_l$ with respect to $x_i$ is denoted as $w_{i,l}$. 
We leverage these sampling points to measure the importance score $I_l$ of the voxel $\mathbf{v}_l$ according to
\begin{equation}\label{eq:importance}
I_l = \sum_i \mathbf{1} \left\{\mathbf{x}_i \in \iota(\mathbf{v}_l)\right\} \cdot w_{i,l}\cdot T_i \cdot \alpha_i,
\end{equation}
where $\mathbf{1}{\{\cdot\}}$ denotes the indicator function and $\iota(\mathbf{v}_l) = \{\mathbf{v}_l + \delta :\delta \in [-1, 1]^3\}$, and $T_i$ and $\alpha_i$ are defined in Eq.~\ref{eq:alpha_t}.

\begin{table}[t]   
    \centering
    \small
    \caption{
    {Importance analysis.}
    }
    \vspace{-0.5mm}
    \begin{tabular}{rccccccc}
        \toprule
        x\% of voxels & 1\%    & 10\%      & 17\%  & 20\% & 100\% \\
        y\% of importance  & 17\%   & 66\%  & 83\% &  86\% & 100\%\\
        \bottomrule
    \end{tabular}
    \label{tab:ablation percent}
    \vspace{-1mm}
\end{table}

In practice, we cast a large set of rays in the source grid model $\mathcal{V}$ across a wide range of views and estimate the importance scores of all voxels as in Eq.~\ref{eq:importance}. The statistic analysis in Table~\ref{tab:ablation percent} shows that over \textbf{83\%} of the overall importance is contributed by only \textbf{17\%} of total voxels. Voxels with large scores form a small proportion. We introduce a rate to denote the percentage of voxels sorted in decreasing scores. As shown in Table \ref{tab:ablation percent}, saving a minimal amount of important voxels is a simple and effective strategy to retain model ability. We thus store a compressed grid and a tiny set of important voxels, reaching a significantly smaller storage compared to the original grid. While the rendering quality is still less satisfied only with nonuniform compression, we therefore propose a parameterized module to further preserve subtle details.

\subsection{Neural Codebook for Details Recovery}
When performing downsampling on grids, many high-frequency details encoded in models are lost. We introduce a new module, namely Neural Codebook (NCB), to encode the model-specific details effectively via a shallow network. The network input and output represent the same type of features with an identical shape, thus it is a nature way to use a residual connection to ease information propagation. In addition, point-wise manipulation may omit contextual information existing in spatial locations of grid models and the model-specific details are usually position-aware, thus embedding position information is necessary. To this end, we design a position modulated network. The design is inspired from style injection in \cite{styleGAN}. 

Formally, assume that we have a pair of grid models $(\mathcal{V},\widetilde{\mathcal{V}})$ where $\widetilde{\mathcal{V}} = \{\widetilde{\mathbf{V}}_{\sigma}, \widetilde{\mathbf{V}}_c\}$ denotes the model with the same dimensions to the source grid $\mathcal{V} = \{\mathbf{V}_{\sigma},\mathbf{V}_c\}$ after upsampling. Neural Codebook is realized by learning a mapping function between the grid pairs, i.e.,
\begin{equation}\label{eq:nc}
\widetilde{\mathbf{V}}_{\sigma}^r, \widetilde{\mathbf{V}}_{c}^r = f_r(\widetilde{\mathbf{V}}_{\sigma}, \widetilde{\mathbf{V}}_{c}, \mathbf{P}).
\end{equation}
where $f_r$ is instantiated with the architecture shown in Fig.~\ref{fig:NCB framework}(d). The output grid model $\widetilde{\mathcal{V}}^r = \{\widetilde{\mathbf{V}}_{\sigma}^r, \widetilde{\mathbf{V}}_{c}^r\}$ is generated by the refinement. 
Specially, we first lift the coordinates to a higher-dimensional embedding. Each voxel $\mathbf{v}$ in the grid is encoded by a high-dimensional mapping $\gamma$ with a set of sinusoidal functions \cite{FourierFeature}:
\begin{equation}\label{eq:sin}
\begin{aligned}
\gamma(\mathbf{v}) = [\cos(2\pi \mathbf{a}^{\top}_1\mathbf{v}), \sin(2\pi \mathbf{a}^{\top}_1\mathbf{v}),\cdots,  \\
\cos(2\pi \mathbf{a}^{\top}_L\mathbf{v}), \sin(2\pi\mathbf{a}^{\top}_L\mathbf{v})],
\end{aligned}
\end{equation}
where each entry of the basis $\mathbf{A} = [\mathbf{a}_1, \cdots, \mathbf{a}_L]$ is sampled from a standard normal distribution. The output position embedding is denoted by $\mathbf{P}$. 

\begin{table}[tbp]
    \caption{Comparison with real-time inference methods on Synthetic-NeRF.}
    \label{tab:results}
    \centering
    \bgroup
    \small
    \vspace{-1mm}
    \begin{tabular}{l@{\hskip 4pt}@{\hskip 5.5pt}|c@{\hskip 5.5pt}@{\hskip 5.5pt}c@{\hskip 5.5pt}c@{\hskip 5.5pt}c@{\hskip 5.5pt}}
    \hline
    & PSNR $\uparrow$    & Deploy Time$\downarrow$ & Test Time$\downarrow$  & Size$\downarrow$ \\
     & (dB)&  (s)   &  (ms)  &  (MB)\\
    \hline
    \hline
    NeRF\cite{NeRF}  & 31.01  & - &  3000 &  5      \\

    \hline
    \multicolumn{2}{l}{ Real-Time Inference Methods} \\
    \hline
    FastNeRF\cite{fastnerf}  & 29.97   &  -  &  4.2  &  -         \\
    SNeRG\cite{SNeRG}         & 30.38  & -  &  12   &   86          \\
    KiloNeRF\cite{kilonerf}   & 31.00  &  5.67 & 11  & 153     \\
    AutoInt\cite{DavidBLindell2020AutoIntAI}       & 26.83 &7.65 & 386  &14\\
    DIVeR \cite{LiwenWu2021DIVeRRA}          & 32.12  & 1.28  & 11.1  & 68  \\
    PlenOctrees\cite{PlenOctrees}   & 31.71  & 3.97  &  3   & 1976       \\
    \hspace{0.2cm} \textsuperscript{$\llcorner$} compressed   & 29.60  & -   &  1.5   & 300           \\
    \hline
    Plenoxels \cite{yu2021plenoxels} & 31.71     &  1.36   &  41   &  779      \\

    Ours  & 31.45  &  1.39  &  5.5 &  18.4     \\
    \hline
    \end{tabular}
    \egroup
\end{table}

We then leverage adaptive layer normalization (AdaLN) to modulate the activation of each fully-connected (FC) layer through a learned transformation on $\mathbf{P}$. The output of the transformation is denoted as $\mathbf{y} = \left(y_b, y_s\right)$. Let $\mathbf{u}$ denote the activation of the FC layer. The AdaLN is formulated as:
\begin{equation}
    \text{AdaLN}(\mathbf{u}, \mathbf{y}) = y_s \cdot \frac{\mathbf{u} - \mu\left(\mathbf{u}\right)}{\nu\left(\mathbf{u}\right)} + y_b,
\end{equation}
where $\mu\left(\right)$ and $\nu\left(\right)$ respectively denote the mean and variance operations across channels. Specially, two modal information are fed into the module. As the correlation between them is expected to be employed, we design the architecture with the property of sharing features at lower layers adjacent to inputs and learning features specialized to respective targets when reaching at higher layers. 
Detailed architecture configuration and training strategies can be found in supplemental material.

The parameters of NCB are model-specific, i.e. preserving detailed properties for each scene model. The network is trained by minimizing the L1 loss, 
\begin{equation}\label{eq:loss}
 \mathcal{L} = \lambda_c\left|\widetilde{\mathbf{V}}_c^r - \mathbf{V}_c\right| + \lambda_\sigma \left|\widetilde{\mathbf{V}}_\sigma^r - \mathbf{V}_\sigma\right|.
\end{equation}

The module optimization is fairly fast, typically costing about 20 minutes in a single GPU (NVIDIA A100). It is possible to accelerate training with a
tailored training schedule, or train a codebook with a shared stem and multi-head
for different scenes, which we leave as future work.
The grid model can be efficiently restored through one-pass forward propagation during the decompression phase, which does not hinder rendering efficiency given novel views.

%% file: exp.tex
\begin{figure}[t!]
    \centering
    \includegraphics[width=0.98\linewidth]{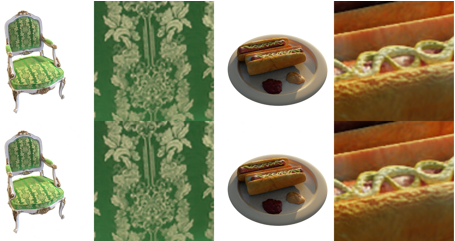}
    \caption{Visual comparison. \textbf{Top: }  Plenoxels  vs \textbf{Bottom:} Ours.}
    \label{fig:Nerf synthetic result}  
\end{figure}
\begin{table}[t]
    \caption{
    Comparison on forward facing scenes.
    }
    \vspace{-0.5mm}
    \label{tab:llff table results}
    \centering
    \small
    \bgroup
    \begin{tabular}{l@{\hskip 12pt}@{\hskip 5.5pt}|c@{\hskip 6.5pt}c@{\hskip 6.5pt}c@{\hskip 6.5pt}}
    \hline
    & PSNR $\uparrow$  & Test Time$\downarrow$  & Size$\downarrow$ \\
     & (dB) &  (ms)  &  (MB)\\
    \hline
    \hline
    NeRF\cite{NeRF}  & 26.50   &  3000 &  13.7 \\

    Plenoxels\cite{yu2021plenoxels} & 26.29     &  41  &  2575     \\
    Ours  & 26.03    &  5.5   &  55.9    \\
    \hline
    \end{tabular}
    \egroup
\end{table}

\section{Experiments}

\subsection{Datasets}

\textbf{Synthetic-NeRF}\cite{NeRF} composed of 8 synthetic objects. Each scene has 100 training views and 200 testing views at 800x800 resolution, with precisely known camera pose.

\noindent\textbf{LLFF} \cite{mildenhall2019local} contains real-world images at resolution $1008\times756$ in a forward facing setting. Each scene contains images from 20 to 60. Camera pose is estimated by COLMAP.

\subsection{Results} 

\textbf{Synthetic Scenes.} We compare NCB with other real-time inference methods on multiple metrics including rendering quality (PSNR), deployment time (in minutes), test time (inference cost given views in ms) and storage size, and list the performance of original NeRF for reference. The results are shown in Table~\ref{tab:results}. 
Our method shows significant advantage on storage cost compared to these methods, specially achieving over $40\times$ reduction over the direct counterpart Plenoxels with minimal quality loss (0.26 dB). We also provide visual comparison between rendering results of Plenoxels and ours in Fig.~\ref{fig:Nerf synthetic result}, showing negligible visual difference between the original and restored models. Moreover, compared to other baselines, our method can also achieve competitive performance on rendering quality, deployment and inference cost, meanwhile realizing a low storage cost.

\begin{figure}[t!]
    \centering
    \includegraphics[width=0.92\linewidth]{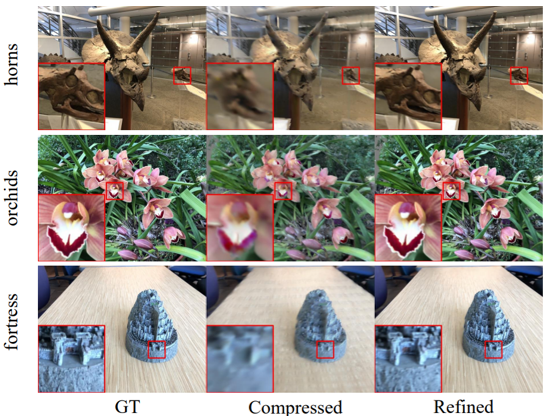}
    \caption{Visual comparison between directly compressed model and recovered model with NCB on LLFF dataset. }
        \label{fig:LLFF result}  
\end{figure}

\textbf{Forward-facing Scenes}. Our method also supports unbounded forward-facing scenes. Following the setting in \cite{NeRF}, We use 7/8 data as the training set, and the left 1/8 data for test. The quantitative results are shown in Table \ref{tab:llff table results}, and the visual results are shown in Figure \ref{fig:LLFF result}. Our method can also achieve a over $40\times$ reduction rate with comparable rendering quality and better inference performance.

\begin{table}[tp]
    \caption{
    {Ablation of saving with different percentage of most important voxels.}
    }
    \label{tab:ablation percent}
    \centering
    \small
    \vspace{-2mm}
    \begin{tabular}{r|cccc}
        \toprule
         & PSNR$\uparrow$ & SSIM$\uparrow$ & LPIPS$\downarrow$ & Add Size$\downarrow$  \\
        \midrule
        0    $\%$   & 31.02 & 0.950  & 0.068    & -    \\
        2.5  $\%$   & 31.35 & 0.954  & 0.062    & 2.6 MB \\
        5.0  $\%$   & 31.45 & 0.943  & 0.058    & 5.2 MB \\
        7.5  $\%$   & 31.50 & 0.956  & 0.056    & 7.6 MB \\
        10   $\%$   & 31.54 & 0.957  & 0.055    & 10.0 MB\\
        \bottomrule
    \end{tabular}
\end{table}

\subsection{Ablation Study}
\textbf{Non-uniform Compression.} 
We conduct the experiments with different downsampling scales on color features and density to estimate their effects on rendering quality. 
The results in Table ~\ref{tab:Nonuniform downsample} show that density contributes more to preserving quality while color features occupy much more storage cost. Therefore we use a smaller scale on color features (1/4 by default) and a moderate scale on density (1/2 by default). 

We also conduct experiments on the effect of saving  different percentage of most important voxels for rendering quality and storage addition in Table \ref{tab:ablation percent}. We can observe that saving with a small fraction can bring obvious benefit on rendering quality with acceptable storage size.

\begin{table}[tbp]
    \caption{
        Ablation of inference acceleration for rendering a 800x800 image on Synthetic-NeRF.  }
    \label{tab:inference accleration}
    \vspace{-2mm}
    \centering
    \begin{tabular}{l|cr}
        \toprule
        &PSNR$\uparrow$ & Test Time$\downarrow$  \\
        \midrule
        Baseline             & 31.58  & 41.8 ms         \\
        + Combine thread & 31.58  & 21.6 ms        \\
        + Larger step        & 31.50  & 12.1 ms   \\
        + Early termination  & 31.45  & 7.8 ms   \\
        + Prefetch neighbor  & 31.45  & 5.5 ms   \\
        \bottomrule
    \end{tabular}
\end{table}

\begin{table}[tp]
    \caption{
        Deployment time analysis for Plenoxels and ours.
    }
    \label{tab:inference time analysis}
    \centering
    \begin{tabular}{l|rr}
        \toprule
                        &Plenoxels & Ours  \\
        \midrule
        Disk IO \& Decode     & 1257 ms  & 97 ms         \\
        Host-device transfer & 110 ms   & 52 ms        \\
        Trilinear upsample   & -        & 30 ms   \\
        NCB inference        & -        & 1219 ms   \\
        \midrule
        Sum                  & 1367 ms   & 1398 ms \\
        \bottomrule
    \end{tabular}

\end{table}

\textbf{Neural Codebook.} 
We further investigate the benefit of neural codebooks by comparing the results with and without NCB, and  use GT for reference. As shown in Figure \ref{fig:LLFF result}, the module is capable of preserving high-frequency details which are lost in the model via directly compression. We also conduct the experiments by learning two networks independently for volume density and color features which achieves 31.38 dB in PSNR (0.07 dB drop) with a larger number of parameters.

\textbf{Inference \& Deployment Time Analysis.} 
We show the ablation in Table~\ref{tab:inference accleration} by adding acceleration strategies, to estimate their effect on speed and PSNR. The strategies can boost inference efficiency obviously with minimal quality drop (0.1 dB). We further compare the deployment time with original Plenoxels through a step-by-step analysis in Table~\ref{tab:inference time analysis}. Plenoxels' disk io time is much longer due to large model size. Our framework can still achieve comparable deployment time, although it need additional decompression procedure.

%% file: supp.tex
\twocolumn[
\centering
\Large
\textbf{Compact Real-time Radiance Fields with Neural Codebook} \\
\vspace{0.5em}Appendix \\
\vspace{1.0em}
] 

\begin{figure}[tbp]
    \centering
    \includegraphics[width=\linewidth]{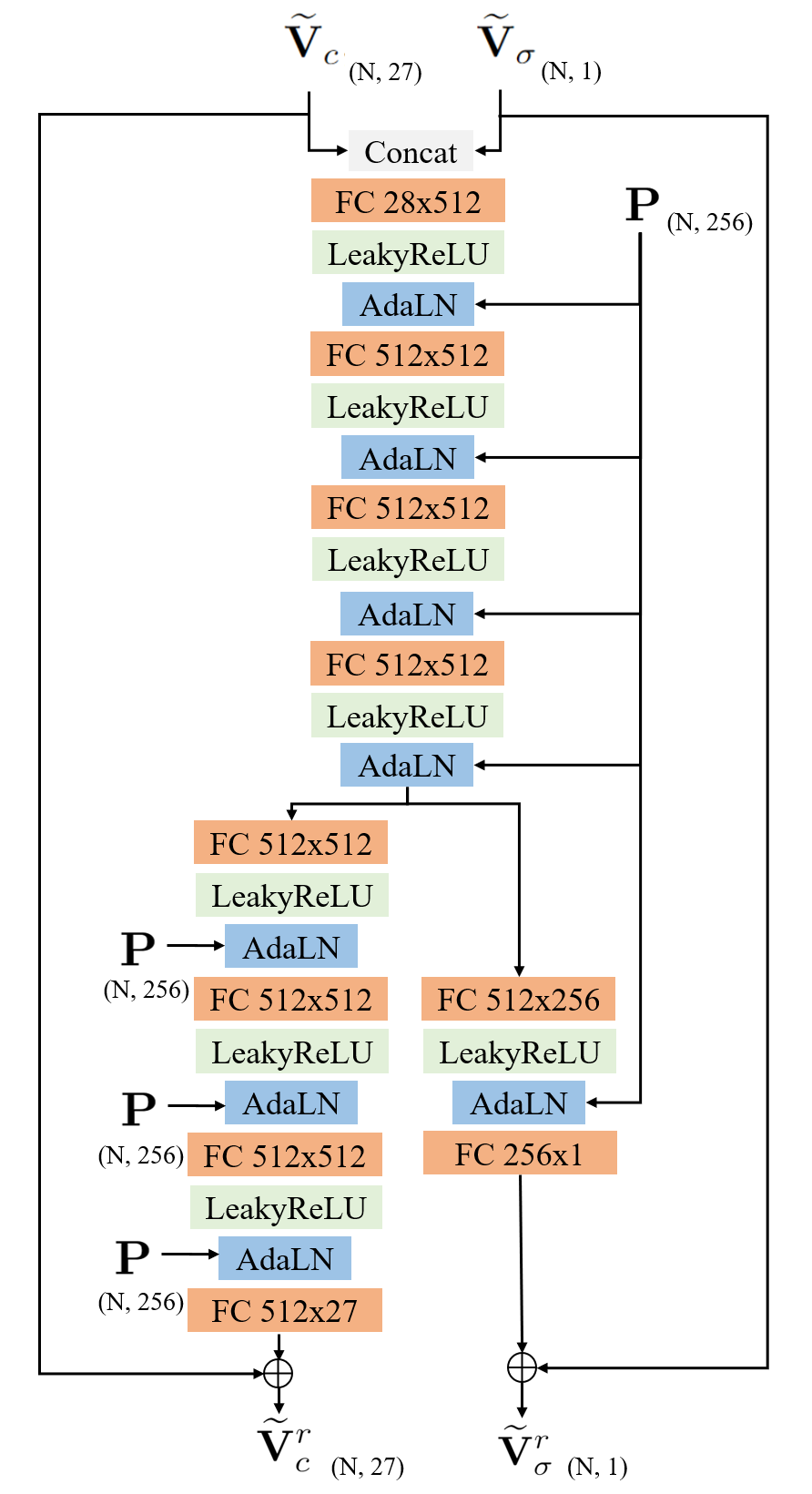}
    \caption{Detailed structure of Neural Codebook}
    \label{fig:network detail}
\end{figure}

\appendix

\section{Implementation Details}

\textbf{Nonuniform compression.} We treat $\mathbf{V}_{\sigma}$ and $\mathbf{V}_{c}$ as dense voxel grid and directly apply trilinear downsampling. We follow the sparsification operation in \cite{yu2021plenoxels} pipeline to re-sparsify the downsapled voxel grids, and keep the full-resolution mask for future decompression. Finally all tensors are covert to float 16 and compressed with zlib \cite{zlib} for saving. During decompression phase, we trilinear upsample $\widehat{\mathbf{V}}_{\sigma}$ and $\widehat{\mathbf{V}}_c$ to get the full-size voxel grid and sparsify it with the saved full-resolution mask. 

The data pointers (both used in importance map and sparse array storage) are converted to a binary mask indicating if a voxel in the grid is occupied, and the conversation is reversible to derive data pointers from the mask. A $4$-byte for saving a data pointer consequentially reduces to $1$-bit.

\noindent\textbf{Neural Codebook.} We train separate Neural Codebooks (NCB) for each scene. Detailed architecture can be found in the following section.
The network is trained using Adam\cite{kingma2014adam} optimizer with $\beta_1$ is 0.9 and $\beta_2$ is 0.999, we also apply a weight decay of 1e-5. The learning rate is set as 5e-3  at first and then is multiply 0.3 for every 5,000 iterations. We train a total of 20,000 iterations with a batch size of 100,000 voxels. After the training phase, the NCB is then cast to float16. All test experiments and storage sizes are reported using the float16 model.

\section{Network Structure}
The configuration of Neural Codebook architecture is shown in Figure \ref{fig:network detail}. LeakyReLU \cite{LeakRelu} is used as layer activation. The color features $\widetilde{\mathbf{V}}_c$ and volume density $\widetilde{\mathbf{V}}_\sigma$ share network layers at lower level adjacent to input and learn modality-specific features through two separate branches at higher level for the refined model $\{\widetilde{\mathbf{V}}^r_c, \widetilde{\mathbf{V}}^r_\sigma\}$, which is optimized by minimizing its difference to the original grid model. The skip connection enables the basic information stored in the compressed model pass through, encouraging the network parameters to capture high-frequency details supplement to the compressed model. By the consideration of higher dimension of color information, the branch corresponding to color information is slightly deeper (with two additional layers) and wider than the density branch. 

\section{Inference Acceleration}
The volume rendering implementation is inherited from Plenoxels \cite{yu2021plenoxels}. We introduce several strategies on both CUDA rendering kernel and rendering strategy to achieve inference speedup. 1) In the original Plenoxels' CUDA kernel, one ray is handled by a full CUDA warp (32 threads) where each thread processes one dimension in SH coefficients in order to take advantage of memory coalescing. We found that such operation would incur large additional computational cost on ray marching which is even more than the benefit of memory coalescing during inference. We instead simplify the procedure to launch one kernel for each ray. 2) The sampling step is twice as the original setting which can decrease the overall sample points. 3) We apply the standard early termination strategy on ray marching. The ray marching would be stopped when the accumulated transmittance is lower than $1\%$. 4) Both SH coefficients and density information are stored in a sparse 3D array with a full-scale mask.  As the SH coefficients of each point is stored in a a contiguous memory space, we only need the data pointer to access the first dimension of SH coefficients and use a fixed stride to prefetch the memory address for the rest dimension, eventually achieving speed-up on the trilinear interpolation kernel.

\section{Hardware Details}
All the experiments ran in a cloud virtual machine, which is equipped with a Intel Xeon Platinum 8369B and 8x NVIDIA A100. As the storage device is undisclosed in the machine, we benchmark the disk with the I/O tester in \cite{fio} for a fair comparison. The average sequential read speed is 540 MB/s and the average random read speed is 375 MB/s.

\section{Results for Each Scene}
\begin{figure*}[b]
    \centering
    \includegraphics[width=0.7\linewidth]{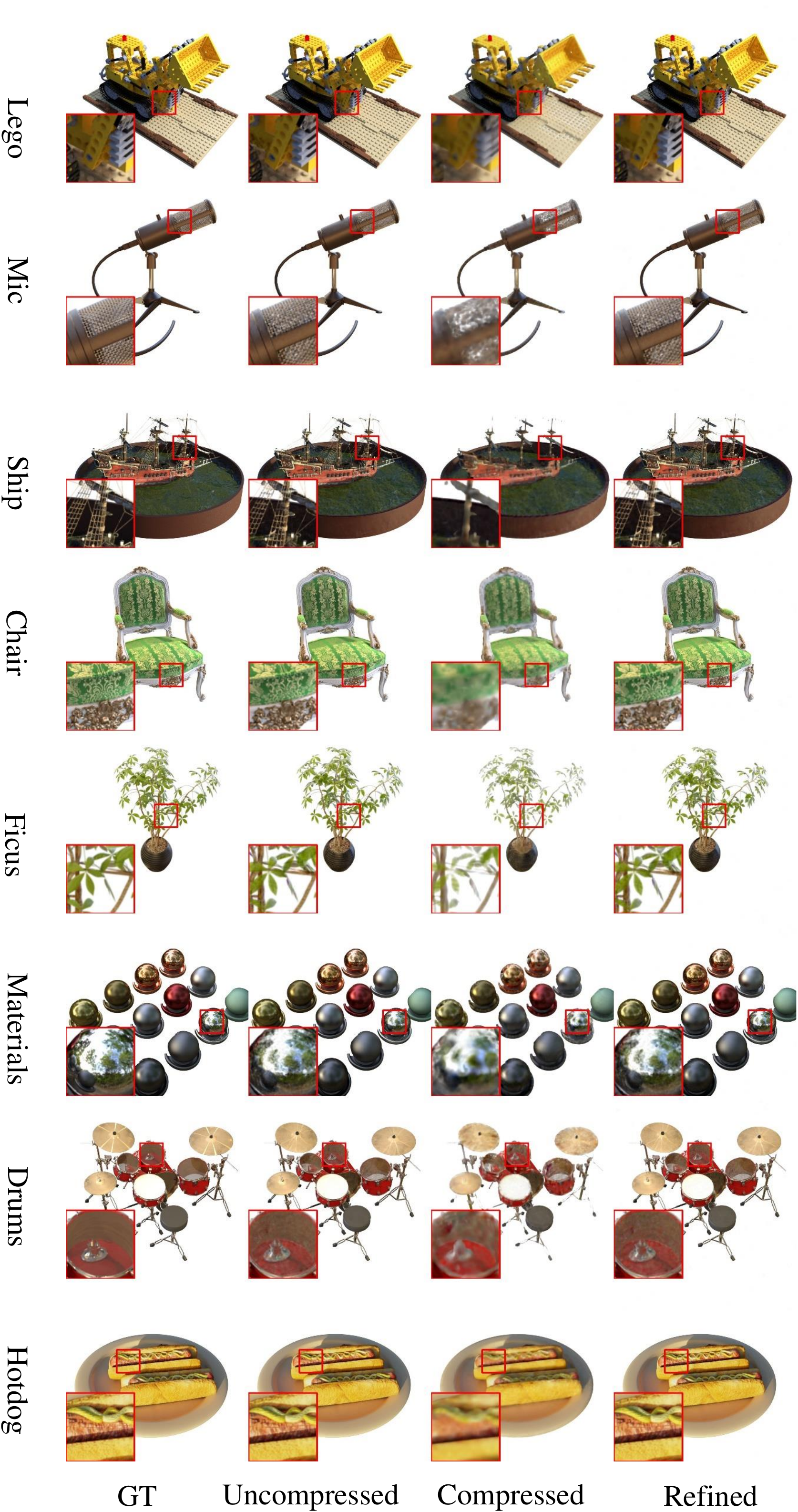}  
    \caption{Qualitative comparison on Synthetic-NeRF dataset.}
    \label{fig:syn visual}
\end{figure*}

\begin{figure*}[b]
    \centering
    \includegraphics[width=0.8\linewidth]{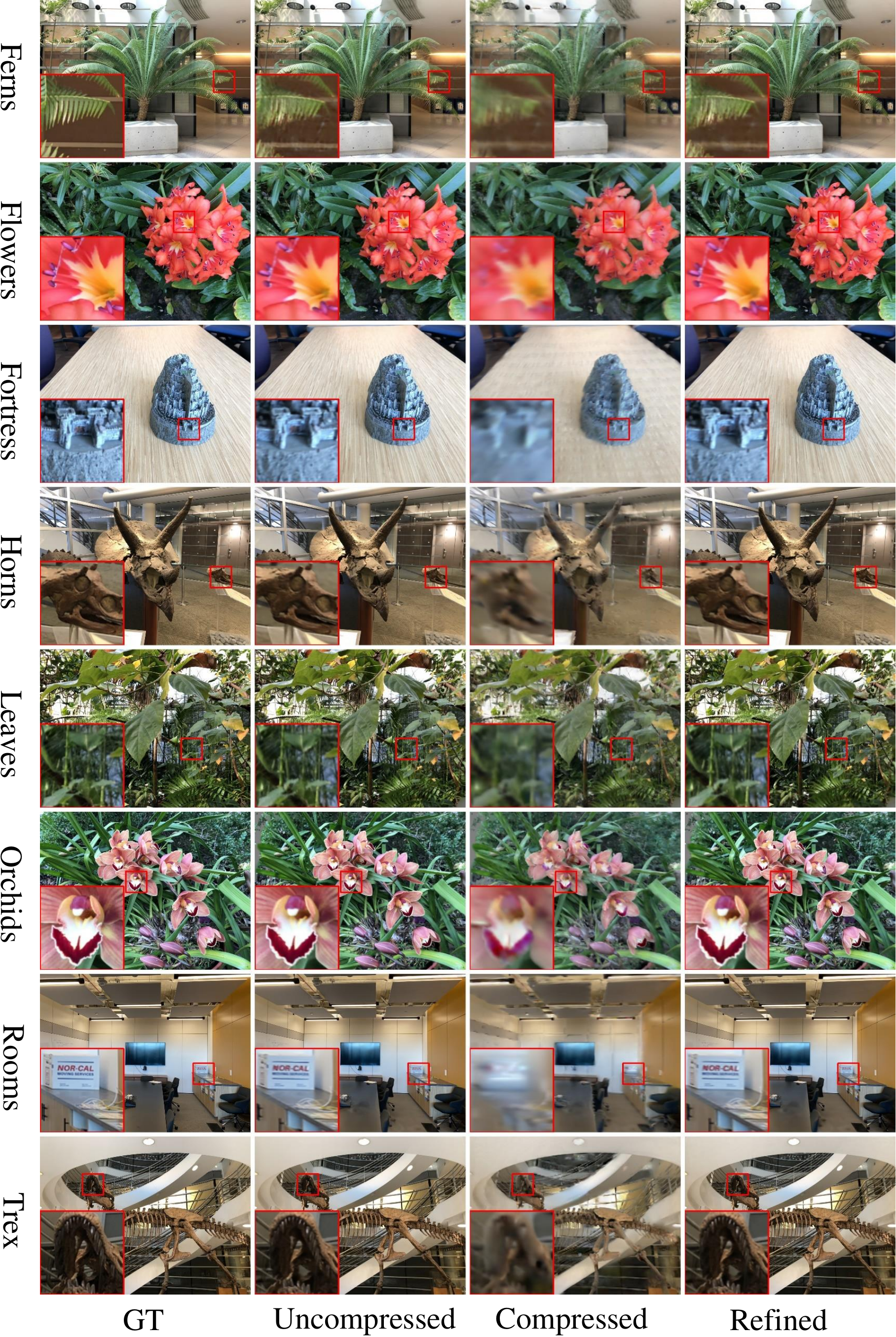}
    \caption{Qualitative comparison on LLFF dataset.}
    \label{fig:llff visual}
\end{figure*}

Table \ref{tab:results} also provides the performance on SSIM and LPIPS besides PSNR. Table \ref{tab:fullsynthetic}  and Table \ref{tab:fullforward}  show full results for each scene on the Synthetic-NeRF \cite{NeRF} dataset and the LLFF \cite{mildenhall2019local} dataset, respectively, demonstrating that compared to Plenoxels our method can achieve comparable rendering quality with significant advantage on storage overhead. 
We further present qualitative comparison among Plenoxels (i.e., Uncompressed), directly compressed model, and refined by using Neural Codebook, by selecting a random view for every scene in Figure \ref{fig:syn visual} and \ref{fig:llff visual}. We can clearly observe that NCB is capable of preserving high-frequency details supplement to the loss in compressed model.

\begin{table*}[h]
    \centering
    \bgroup

    \begin{tabular}{l@{\hskip 15pt}@{\hskip 5.5pt}|c@{\hskip 5.5pt}c@{\hskip 5.5pt}c@{\hskip 5.5pt}|@{\hskip 5.5pt}c@{\hskip 5.5pt}c@{\hskip 5.5pt}c@{\hskip 5.5pt}}
    \hline
    ~ & PSNR$\uparrow$ & SSIM$\uparrow$ & LPIPS$\downarrow$  & Deploy Time$\downarrow$ & Test Time$\downarrow$ & Storage Size$\downarrow$ \\
    \hline
    \hline
    NeRF\cite{NeRF}  & 31.01 & 0.947 & 0.081  & - &  3000 ms &  5 MB     \\
    \hline
    \hline
    \multicolumn{6}{@{}l}{ Real-Time Inference Methods} \\
    \hline
    FastNeRF\cite{fastnerf}  & 29.97 & 0.941 & 0.053  &  -  &  4.2 ms &  -         \\
    SNeRG\cite{SNeRG}         & 30.38 & 0.950 & 0.050 & -  &  12 ms  &   86 MB         \\
    KiloNeRF\cite{kilonerf}   & 31.00 & 0.950  & 0.030 &  5.67 s & 11 ms & 153 MB    \\
    AutoInt\cite{DavidBLindell2020AutoIntAI}         & 26.83 & 0.926&0.151&7.65 s& 386 ms &14.8MB\\
    DIVeR \cite{LiwenWu2021DIVeRRA}          & 32.12 & 0.958 & 0.033 & 1.28 s & 11.1 ms & 68 MB \\
    PlenOctrees\cite{PlenOctrees}   & 31.71 & 0.958 & 0.053 & 3.97 s  &  3 ms  & 1976 MB      \\
    \hspace{0.2cm} \textsuperscript{$\llcorner$} compressed   & 29.60 & - & - & -   &  1.5 ms  & 300 MB          \\
    
    \hline
    Plenoxels \cite{yu2021plenoxels} & 31.71 & 0.958 & 0.049    &  1.36 s   &  41 ms  &  779 MB     \\

    Ours  & 31.45 & 0.955 & 0.058   &  1.39 s  &  5.5 ms  &  18.4 MB    \\
    \hline

    \end{tabular}
    \egroup
    \caption{
    Comparison with other real time methods in synthetic-NeRF.}
    \label{tab:results}
\end{table*}

\renewcommand{\tabcolsep}{3pt}
\begin{table*}[ht!]
  \centering
  \caption{Full results on synthetic-NeRF dataset.}
  \begin{tabular}{llcccccccclc}
    \multicolumn{12}{c}{PSNR $\uparrow$} \\
    \toprule
     & & Chair & Drums & Ficus & Hotdog & Lego & Materials & Mic & Ship & & Mean \\ \cmidrule(){1-1} \cmidrule(){3-10} \cmidrule(){12-12}
Ours & & 33.51 & 25.24 & 31.74 & 36.13 & 33.50 & 29.02 & 33.11 & 29.36 & & 31.45 \\
Plenoxels & & 33.98 & 25.35 & 31.83 & 36.43 & 34.10 & 29.14 & 33.26 & 29.62 & & 31.71 \\
    \bottomrule
 & & & & & & & & & & & \\ 
    \multicolumn{12}{c}{SSIM $\uparrow$} \\
    \toprule
     & & Chair & Drums & Ficus & Hotdog & Lego & Materials & Mic & Ship & & Mean \\ \cmidrule(){1-1} \cmidrule(){3-10} \cmidrule(){12-12}
Ours & & 0.973 & 0.929 & 0.975 & 0.978 & 0.970 & 0.947 & 0.984 & 0.887 & & 0.955 \\
Plenoxels\cite{yu2021plenoxels} & & 0.977 & 0.933 & 0.976 & 0.980 & 0.975 & 0.949 & 0.985 & 0.890 & & 0.958 \\
    \bottomrule
 & & & & & & & & & & & \\ 
    \multicolumn{12}{c}{LPIPS $\downarrow$} \\
    \toprule
     & & Chair & Drums & Ficus & Hotdog & Lego & Materials & Mic & Ship & & Mean \\ \cmidrule(){1-1} \cmidrule(){3-10} \cmidrule(){12-12}
Ours & & 0.040 & 0.079 & 0.030 & 0.044 & 0.039 & 0.066 & 0.020 & 0.148 & & 0.058 \\
Plenoxels & & 0.031 & 0.067 & 0.026 & 0.037 & 0.028 & 0.057 & 0.015 & 0.134 & & 0.049 \\
    \bottomrule
 & & & & & & & & & & & \\ 
    \multicolumn{12}{c}{Storage $\downarrow$} \\
    \toprule
     & & Chair & Drums & Ficus & Hotdog & Lego & Materials & Mic & Ship & & Mean \\ \cmidrule(){1-1} \cmidrule(){3-10} \cmidrule(){12-12}
Ours & & 15.1MB & 14.4MB & 11.9MB & 19.4MB & 19.7MB & 15.9MB & 10.8MB & 39.7MB & & 18.4MB \\
Plenoxels\cite{yu2021plenoxels} & & 701MB & 677MB & 623MB & 814MB & 811MB & 713MB & 595MB & 1300MB & & 779MB \\
    \bottomrule
  \end{tabular}
\newline
  \label{tab:fullsynthetic}
\end{table*}

\renewcommand{\tabcolsep}{3pt}
\begin{table*}[ht!]
  \centering
  \caption{Full results on LLFF dataset.}
  \begin{tabular}{llcccccccclc}
    \multicolumn{12}{c}{PSNR $\uparrow$} \\
    \toprule
     & & Fern & Flower & Fortress & Horns & Leaves & Orchids & Room & T-Rex & & Mean \\ \cmidrule(){1-1} \cmidrule(){3-10} \cmidrule(){12-12}
Ours & & 25.20 & 27.50 & 30.80 & 27.33 & 21.01 & 20.28 & 29.91 & 26.20 & & 26.03 \\
Plenoxels & & 25.46 & 27.83 & 31.09 & 27.58 & 21.41 & 20.24 & 30.22 & 26.48 & & 26.29 \\
    \bottomrule
 & & & & & & & & & & & \\ 
    \multicolumn{12}{c}{SSIM $\uparrow$} \\
    \toprule
     & & Fern & Flower & Fortress & Horns & Leaves & Orchids & Room & T-Rex & & Mean \\ \cmidrule(){1-1} \cmidrule(){3-10} \cmidrule(){12-12}
Ours & & 0.821 & 0.843 & 0.874 & 0.847 & 0.717 & 0.680 & 0.933 & 0.882 & & 0.824 \\
Plenoxels & & 0.832 & 0.862 & 0.885 & 0.857 & 0.760 & 0.687 & 0.937 & 0.890 & & 0.839 \\
    \bottomrule
 & & & & & & & & & & & \\ 
    \multicolumn{12}{c}{LPIPS $\downarrow$} \\
    \toprule
     & & Fern & Flower & Fortress & Horns & Leaves & Orchids & Room & T-Rex & & Mean \\ \cmidrule(){1-1} \cmidrule(){3-10} \cmidrule(){12-12}
Ours & & 0.248 & 0.215 & 0.201 & 0.250 & 0.257 & 0.259 & 0.207 & 0.257 & & 0.237 \\
Plenoxels & & 0.224 & 0.179 & 0.180 & 0.231 & 0.198 & 0.242 & 0.192 & 0.238 & & 0.210 \\
    \bottomrule
     & & & & & & & & & & & \\ 
    \multicolumn{12}{c}{Storage $\downarrow$} \\
    \toprule
     & & Fern & Flower & Fortress & Horns & Leaves & Orchids & Room & T-Rex & & Mean \\ \cmidrule(){1-1} \cmidrule(){3-10} \cmidrule(){12-12}
Ours & & 53.8 MB & 56.7 MB & 61.5 MB & 69.5 MB & 44.6 MB & 64.3 MB & 53.6 MB & 66.5 MB & & 58.9 MB \\
Plenoxels & & 2600 MB & 2500 MB & 2600 MB & 2800 MB & 2100 MB & 2800 MB & 2500 MB & 2700 MB & & 2575 MB \\
    \bottomrule
  \end{tabular}
\newline
  \label{tab:fullforward}
\end{table*}

